\documentclass{article}

\PassOptionsToPackage{numbers, sort&compress}{natbib}

\usepackage[final]{nips_2018}

\usepackage[utf8]{inputenc} 
\usepackage[T1]{fontenc}    
\usepackage{hyperref}       
\usepackage{url}            
\usepackage{booktabs}       
\usepackage{amsfonts}       
\usepackage{nicefrac}       
\usepackage{microtype}      
\usepackage{graphicx}
\usepackage{amsmath}

\newcommand\etc{etc\@ifnextchar.{}{.\@}}
\newcommand\wrt{w.r.t. \@}

\newcommand\ie{i.e., \@}
\newcommand\vs{vs. \@}

\newcommand{\lp}{\left(}
\newcommand{\rp}{\right)}
\newcommand{\rb}{\right]}
\newcommand{\lb}{\left[}

\title{Learning long-range spatial dependencies \\ with horizontal gated recurrent units}

\author{
  Drew Linsley, \hfill Junkyung Kim, \hfill Vijay Veerabadran, \hfill Charlie Windolf, \hfill Thomas Serre\\
  Carney Institute for Brain Science \\
  Department of Cognitive Linguistic \& Psychological Sciences\\
  Brown University \\
  Providence, RI 02912 \\
  \texttt{\{drew\_linsley,junkyung\_kim,vijay\_veerabadran,thomas\_serre\}@brown.edu} \\
}

\begin{document}
\maketitle

\begin{abstract} 
Progress in deep learning has spawned great successes in many engineering applications. As a prime example, convolutional neural networks, a type of feedforward neural networks, are now approaching -- and sometimes even surpassing -- human accuracy on a variety of visual recognition tasks. Here, however, we show that these neural networks and their recent extensions struggle in recognition tasks where co-dependent visual features must be detected over long spatial ranges. We introduce a visual challenge, Pathfinder, and describe a novel recurrent neural network architecture called the horizontal gated recurrent unit (hGRU) to learn intrinsic horizontal connections -- both within and across feature columns. We demonstrate that a single hGRU layer matches or outperforms all tested feedforward hierarchical baselines including state-of-the-art architectures with orders of magnitude more parameters. 
\end{abstract}

\section{Introduction}

\begin{figure}[]
\begin{center}
   \includegraphics[width=1\linewidth]{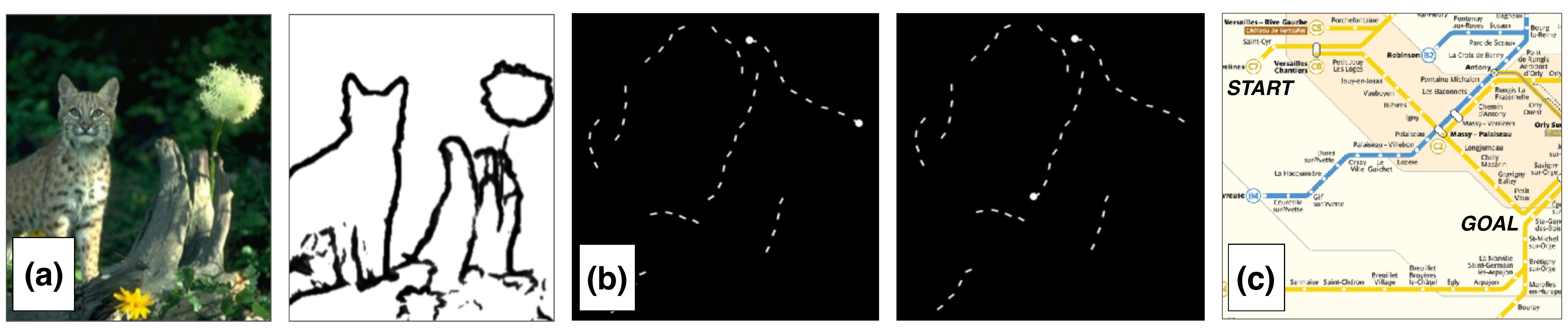}
\end{center}
   \caption{State-of-the-art CNNs excel at detecting contours in natural scenes, but they are strained by a task that requires the detection of long-range spatial dependencies. (a) Representative contour detection performance of a leading neural network architecture~\citep{Xie2015-eh}. (b) Exemplars from the Pathfinder challenge: a task consisting of synthetic images which are parametrically controlled for long-range dependencies. (c) Long-range dependencies similar to those in the Pathfinder challenge are critical for everyday behaviors, such as reading a subway map to navigate a city.}\vspace{-2mm}
\label{fig:task}
\end{figure}

Consider Fig.~\ref{fig:task}a which shows a sample image from a representative segmentation dataset~\citep{Arbelaez2011-lw} (left) and the corresponding contour map produced by a state-of-the-art deep convolutional neural network (CNN)~\citep{Lee2015-mp} (right). Although this task has long been considered challenging because of the need to integrate global contextual information with inherently ambiguous local edge information, modern CNNs are capable to detect contours in natural scenes at a level that rivals that of human observers~\citep{Lee2015-mp, Xie2017-or, Liu2017-ta, Maninis2018-eo, Wang2018-du}. Now, consider Fig.~\ref{fig:task}b which depicts a variant of a visual psychology task referred to as ``Pathfinder''~\citep{Houtkamp2010-te}. Reminiscent of the everyday task of reading a subway map to plan a commute (Fig.~\ref{fig:task}c), the goal in Pathfinder is to determine if two white circles in an image are connected by a path. These images are visually simple compared to natural images like the one shown in Fig.~\ref{fig:task}a, and the task is indeed easy for human observers~\citep{Houtkamp2010-te}. Nonetheless, we will demonstrate that modern CNNs struggle to solve this task. 

Why is it that a CNN can accurately detect contours in a natural scene like Fig.~\ref{fig:task}a but also struggle to integrate paths in the stimuli shown in Fig.~\ref{fig:task}b? In principle, the ability of CNNs to learn such long-range spatial dependencies is limited by their localized receptive fields (RFs) -- hence the need to consider deeper networks because they allow the buildup of larger and more complex RFs. Here, we use a large-scale analysis of CNN performance on the Pathfinder challenge to demonstrate that simply increasing depth in feedforward networks constitutes an inefficient solution to learning the long-range spatial dependencies needed to solve the Pathfinder challenge.



An alternative solution to problems that stress long-range spatial dependencies is provided by biology. The visual cortex contains abundant horizontal connections which mediate non-linear interactions between neurons across distal regions of the visual field~\citep{Stettler2002-hb, Rockland1983-ki}. These intrinsic connections, popularly called ``association fields'', are thought to form the main substrate for mechanisms of contour grouping according to Gestalt principles, by mutually exciting colinear elements while also suppressing clutter elements that do not form extended contours~\citep{Grossberg1985-ui,Field1993-yp,Lesher1993-if,Li2006-ns,Li2008-ac,Zucker2014-mc}. Such ``extra-classical receptive field'' mechanisms, mediated by horizontal connections, allow receptive fields to adaptively ``grow'' without additional processing depth. Building on previous computational neuroscience work~\citep[e.g.,][]{Grossberg1985-ui,Series2003-mg,Zhaoping2011-dn,Shushruth2012-dv,Rubin2015-ws}, our group has recently developed a recurrent network model of classical and extra-classical receptive fields that is constrained by the anatomy and physiology of the visual cortex~\citep{Mely2018-bc}. The model was shown to account for diverse visual illusions providing computational evidence for a novel canonical circuit that is shared across visual modalities. 

Here, we show how this computational neuroscience model can be turned into a modern end-to-end trainable neural network module. We describe an extension of the popular gated recurrent unit (GRU)~\citep{Cho2014-kn}, which we call the horizontal GRU (hGRU). Unlike CNNs, which exhibit a sharp decrease in accuracy for increasingly long paths, we show that the hGRU is highly effective at solving the Pathfinder challenge with just \emph{one layer} and a fraction of the number of parameters and training samples needed by CNNs. We further find that, when trained on natural scenes, the hGRU learns connection patterns that coarsely resemble anatomical patterns of horizontal connectivity found in the visual cortex, and exhibits a detection profile that strongly correlates with human behavior on a classic contour detection task~\citep{Li2002-bq}.


\vspace{-2mm}
\paragraph{Related work}
Much previous work on recurrent neural networks (RNNs) has focused on modeling sequences with learnable gates in the form of long-short term memory (LSTM) units~\citep{Hochreiter1997-gc} or gated recurrent units (GRUs)~\citep{Cho2014-kn}. RNNs have also been extended to learning spatial dependencies in static images with broad applications~\citep{Graves2007-db,Graves2009-hm,Bell2016-kt,Theis2015-wm,Van_den_Oord2016-zt}. In this approach, images are transformed into one-dimensional sequences that are used to train an RNN. In recent years, several approaches have introduced convolutions into RNNs, using the recursive application of convolutional filters as a method for increasing the depth of processing through time on tasks like object recognition and super-resolution without additional parameters~\citep{Liang2015-li,Liao2016-me,Kim2016-pg}. Other groups have constrained these convolutional-RNNs with insights from neuroscience and cognitive science, engineering specific patterns of connectivity between processing layers~\citep{Spoerer2017-ee,Lotter2016-ej,Zamir2016-lr,Nayebi2018-dc}. The proposed hGRU builds on this line of biologically-inspired implementations of RNNs, adding connectivity patterns and circuit mechanisms that are typically found in computational neuroscience models of neural circuits~\citep[e.g.,][]{Grossberg1985-ui,Series2003-mg,Zhaoping2011-dn,Shushruth2012-dv,Rubin2015-ws, Mely2018-bc}.

Another class of models related to our proposed approach is Conditional Random Fields (CRFs), probabilistic models aimed at explicitly capturing associations between nearby features. The connectivity implemented in CRFs is similar to the horizontal connections used in the hGRU, and has been successfully applied as a post-processing stage in visual tasks such as segmentation~\citep{Kokkinos2016-iz,Chen2018-tq} to smooth out and increase the spatial resolution of prediction maps. Recently, such probabilistic methods have been successfully incorporated in a generative vision model shown to break text-based CAPTCHAs~\citep{George2017-ae}. Originally formulated as probabilistic models, CRFs can also be cast as RNNs~\citep{Zheng2015-fo}.  



\section{Horizontal gated recurrent units (hGRUs)}\label{our_model}


\paragraph{Original contextual neural circuit model} 
We begin by referencing the recurrent neural model of contextual interactions developed by~\citet{Mely2018-bc}. Below we adapted the model notations to a computer vision audience. Model units are indexed by their 2D positions $(x, y)$ and feature channel $k$. Neural activity is governed by the following differential equations (see Supp. Material for the full treatment):

\vspace{-3mm}\begin{align}
\begin{split}
\eta \dot{H}^{(1)}_{xyk} + \epsilon^2 H^{(1)}_{xyk}
&=
\lb \xi X_{xyk} - (\alpha H^{(1)}_{xyk} + \mu)\, C^{(1)}_{xyk} \rb_+ \
\label{newx}
\\
\tau \dot{H}^{(2)}_{xyk} + \sigma^2 H^{(2)}_{xyk}
&=
\lb \gamma C^{(2)}_{xyk} \rb_+. \
\end{split}
\end{align}
where
\vspace{-3mm}\begin{align*}
C^{(1)}_{xyk} &= (\textbf{W}^{I} * \textbf{H}^{(2)})_{xyk}\\
C^{(2)}_{xyk} &= (\textbf{W}^{E} * \textbf{H}^{(1)})_{xyk},
\end{align*}





\vspace{-2mm}
Here, $\textbf{X} \in \mathbb{R}^{\textit{W} \times \textit{H} \times \textit{K}}$ is the feedforward drive (\ie neural responses to a stimulus), $\textbf{H}^{(1)} \in \mathbb{R}^{\textit{W} \times \textit{H} \times \textit{K}}$ is the recurrent circuit input, and $\textbf{H}^{(2)} \in \mathbb{R}^{\textit{W} \times \textit{H} \times \textit{K}}$ the recurrent circuit output. Modeling input and output states separately allows for the implementation of a particular form of inhibition known as ``shunting'' (or divisive) inhibition. Unlike the excitation in the model which acts linearly on a unit's input, inhibition acts on a unit's output and hence, regulates the unit response non-linearly (\ie given a fixed amount of inhibition and excitation, inhibition will increase with the unit's activity unlike excitation which will remained constant).

The convolutional kernels $\textbf{W}^{I}, \textbf{W}^{E} \in \mathbb{R}^{\textit{S} \times \textit{S} \times \textit{K} \times \textit{K}}$ describe inhibitory \vs excitatory hypercolumn connectivity (constrained by anatomical data\footnote{There are four separate connectivity patterns in \citep{Mely2018-bc} to describe inhibition \vs excitation and near \vs far interactions between units. We combine these into a separate inhibitory \vs excitatory kernels to simplify notation.}). The scalar parameters $\mu$ and $\alpha$ control linear and quadratic (\ie shunting) inhibition by $\textbf{C}^{(1)} \in \mathbb{R}^{\textit{W} \times \textit{H} \times \textit{K}}$, $\gamma$ scales excitation by $\textbf{C}^{(2)} \in \mathbb{R}^{\textit{W} \times \textit{H} \times \textit{K}}$, and $\xi$ scales the feedforward drive. Activity at each stage is linearly rectified (ReLU) $[\cdot]_+=\max(\cdot,0)$. Finally, $\eta$, $\epsilon$, $\tau$ and $\sigma$ are time constants. To make this model amenable to modern computer vision applications, we set out to develop a version where all parameters could be trained from data. If we let $\eta=\tau$ and $\sigma=\epsilon$ for symmetry and apply Euler's method to Eq.~\ref{newx} with a time step of $\Delta t=\eta/\epsilon^2$, then we obtain the discrete-time equations:

\vspace{-3mm}\begin{align}
H_{xyk}^{(1)}[t]&=\epsilon^{-2}\lb \xi X_{xyk} - (\alpha H_{xyk}^{(1)}[t-1] + \mu)C_{xyk}^{(1)}[t]\rb_+\nonumber\\
H_{xyk}^{(2)}[t]&=\epsilon^{-2}\lb\gamma C_{xyk}^{(2)}[t]\rb_+.
\label{euler}
\end{align}
\vspace{-3mm}

Here, $\cdot[t]$ denotes the approximation at the $t$-th discrete timestep. This results in a trainable convolutional recurrent neural network (RNN) which performs Euler integration of a dynamical system similar to the neural model of~\citep{Mely2018-bc}.

\begin{figure}[t!]
\begin{center}
   \includegraphics[width=1\linewidth]{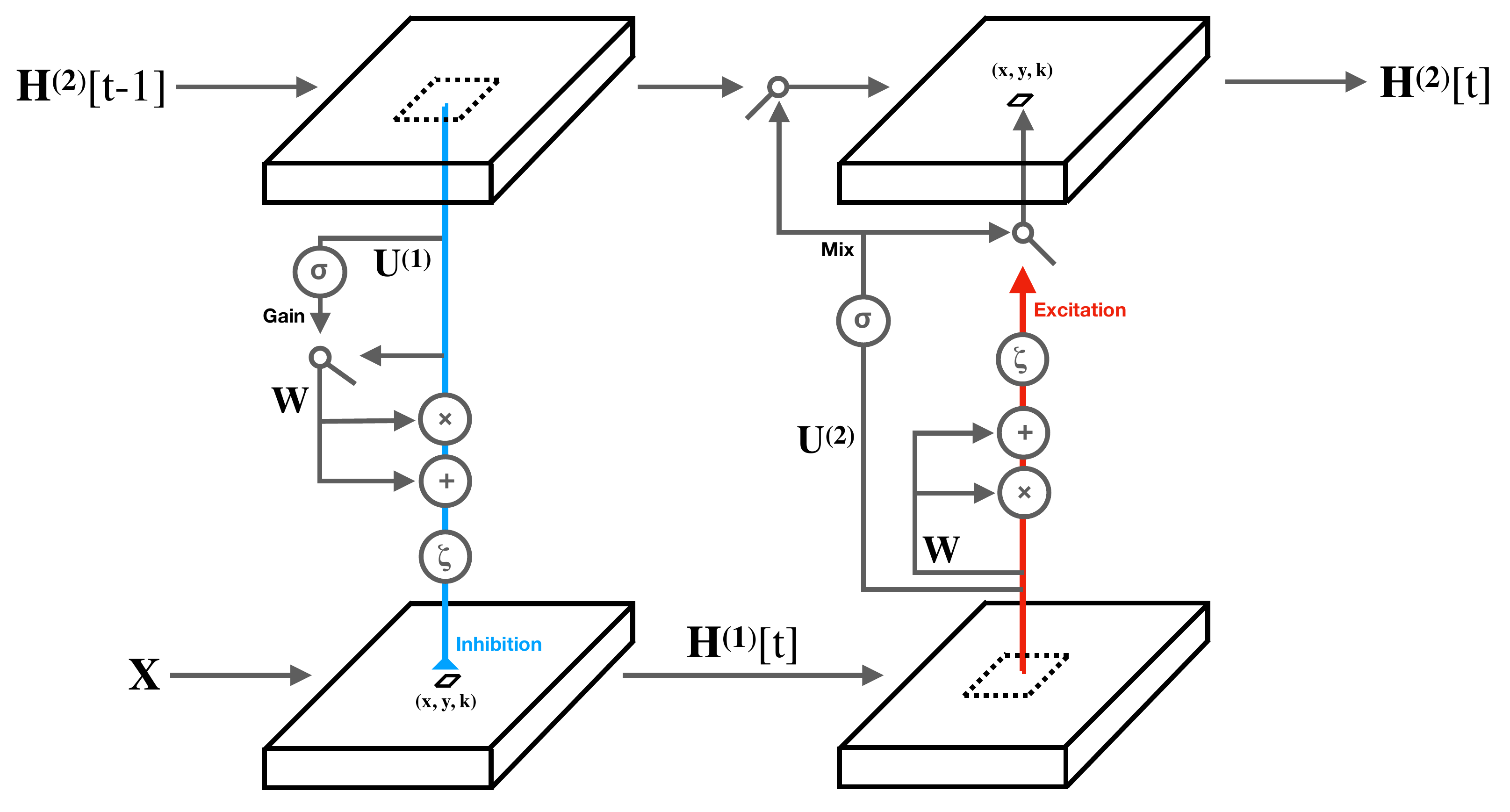}
\end{center}
   \caption{The hGRU circuit. The hGRU can learn highly non-linear interactions between spatially neighboring units in the feedforward drive $\textbf{X}$, which are encoded in its hidden state $\textbf{H}^{(2)}$. This computation involves two stages, which are inspired by a recurrent neural circuit of horizontal connections \citep{Mely2018-bc}. First, the horizontal inhibition (blue) is calculated by applying a gain to $\textbf{H}^{(2)}[t-1]$, and convolving the resulting activity with the kernel $W$ which characterizes these interactions. Linear ($+$ symbol) and quadratic ($\times$ symbol) operations control the convergence of this inhibition onto $\textbf{X}$. Second, the horizontal excitation (red) is computed by convolving $\textbf{H}^{(1)}[t]$ with $W$. Another set of linear and quadratic operations modulate this activity, before it is mixed with the persistent hidden state $\textbf{H}^{(2)}[t-1]$. Note that the excitation computation involves an additional ``peephole'' connection, not depicted here. Small solid-line squares within the hypothetical activities that the circuit operates on denote the unit indexed by 2D position $(x, y)$ and feature channel $k$, whereas dotted-line squares depict the unit's receptive field (a union of both classical and extra-classical definitions) in the previous activity.}
\label{model}
\end{figure}



\vspace{-2mm}
\paragraph{hGRU formulation} We build on Eq.~\ref{euler} to introduce the hGRU -- a model with the ability to learn complex interactions between units via horizontal connections within a single processing layer (Fig.~ \ref{model}). The hGRU extends the derivation from Eq.~\ref{euler} with three modifications that improve the training of the model with gradient descent and its expressiveness\footnote{These modifications involved relaxing several constraints from the original neuroscience model that are less useful for solving the tasks investigated here (see Supp. Material for performance of an hGRU with constrained inhibition and excitation.)}. (i) We introduce learnable gates, borrowed from the gated recurrent unit (GRU) framework (see Supp. Material for the full derivation from Eq.~\ref{euler}). (ii) The hGRU makes the operations for computing $\textbf{H}^{(2)}$ (excitation) symmetric with those of $\textbf{H}^{(1)}$ (inhibition), providing the circuit the ability to learn how to implement linear and quadratic interactions at each of these processing stages. (iii) To control unstable gradients, the hGRU uses a squashing pointwise non-linearity and a learned parameter to globally scale activity at every processing timestep (akin to a constrained version of the recurrent batchnorm \citep{Cooijmans2017-bf}).

In our hGRU implementation, the feedforward drive $\textbf{X}$ corresponds to activity from a preceding convolutional layer. The hGRU encodes spatial dependencies between feedforward units via its (time-varying) hidden states $\textbf{H}^{(1)}$ and $\textbf{H}^{(2)}$. Updates to the hidden states are managed using two activities, referred to as the reset and update ``gates'': $\textbf{G}^{(1)}$ and $\textbf{G}^{(2)}$. These activities are derived from convolutions, denoted by $*$, between the kernels ${\textbf{U}^{(1)}, \textbf{U}^{(2)}} \in \mathbb{R}^{\textit{1} \times \textit{1} \times \textit{K} \times \textit{K}}$ and hidden states $\textbf{H}^{(1)}$ and $\textbf{H}^{(2)}$, shifted by biases  $\textbf{b}^{(1)}, \textbf{b}^{(2)} \in \mathbb{R}^{\textit{1} \times \textit{1} \times \textit{K}}$, respectively. The pointwise non-linearity $\sigma$ is applied to each activity, normalizing them in the range $[0, 1]$. Because these activities are real-valued, we hereafter refer to the reset gate as the ``gain'', and the update gate as the ``mix''.




Horizontal interactions between units are calculated by the kernel $\textbf{W} \in \mathbb{R}^{\textit{S} \times \textit{S} \times \textit{K} \times \textit{K}}$, where $S$ describes the spatial extent of these connections in a single timestep (Fig.~\ref{model}; but see Supp. Material for a version with separate kernels for excitation \vs inhibition, as in Eq.~\ref{euler}). Consistent with computational models of neural circuits (e.g., ~\citep{Grossberg1985-ui,Series2003-mg,Zhaoping2011-dn,Shushruth2012-dv,Rubin2015-ws, Mely2018-bc}), $\textbf{W}$ is constrained to have symmetric weights between channels, such that the weight $W_{x_0 + \Delta x, y_0 + \Delta y, k_1, k_2}$ is equal to the weight $W_{x_0 + \Delta x, y_0 + \Delta y, k_2, k_1}$ where $x_0$ and $y_0$ denote the center of the kernel. This constraint reduces the number of learnable parameters by nearly half \vs a normal convolutional kernel. Hidden states $\textbf{H}^{(1)}$ and $\textbf{H}^{(2)}$ are recomputed via horizontal interactions at every timestep $t \in [0, T]$. We begin by describing computation of $\textbf{H}^{(1)}[t]$: 

\vspace{-5mm}
\begin{align}
    \textbf{G}^{(1)}[t] &= \sigma(\textbf{U}^{(1)} * \textbf{H}^{(2)}[t-1] + \textbf{b}^{(1)})\\
    {C}^{(1)}_{xyk}[t] &= (\textbf{W} *   (\textbf{G}^{(1)}[t] \odot \textbf{H}^{(2)}[t-1]))_{xyk}\\
    {H}^{(1)}_{xyk}[t] &= \zeta({X}_{xyk} - {C}^{(1)}_{xyk}[t](\alpha_{k} {H}^{(2)}_{xyk}[t-1] +\mu_{k}))\label{inhibition}
\end{align}\vspace{-3mm}

Channels in $\textbf{H}^{(2)}[t-1]$ are first modulated by the gain\footnote{GRU gate activities are often a function of a hidden state and $\textbf{X}[t]$. Because the feedforward drive here is constant w.r.t. time, we omit it from these calculations. In practice, its inclusion did not affect performance.} $\textbf{G}^{(1)}[t]$. The resulting activity is convolved with $\textbf{W}$ to compute $\textbf{C}^{(1)}[t]$, which is the horizontal inhibition of the hGRU at this timestep. This inhibition is applied to $\textbf{X}$ via the parameters $\boldsymbol{\mu}$ and $\boldsymbol{\alpha}$, which are $k$-dimensional vectors that respectively scale linear  and quadratic (akin to shunting inhibition described in Eq.~\ref{newx}) terms of the horizontal interaction with $\textbf{X}$. The pointwise $\zeta$ is a hyperbolic tangent that squashes activity into the range $[-1, 1]$ (but see Supp. Material for a hGRU with a rectified linearity). Importantly, in contrast to the original circuit, in this formulation the update to $\textbf{H}^{(1)}[t]$ (Eq.~\ref{inhibition}) is calculated by combining horizontal connection contributions of $\textbf{C}^{(1)}[t]$ with $\textbf{H}^{(2)}[t-1]$ rather than $\textbf{H}^{(1)}[t-1]$, which we found improved learning on the visual tasks explored here.




The updated $\textbf{H}^{(1)}[t]$ is next used to calculate $\textbf{H}^{(2)}[t]$.
\begin{align}\vspace{-5mm}
    {G}^{(2)}_{xyk}[t] &= \sigma((\textbf{U}^{(2)} * \textbf{H}^{(1)}[t])_{xyk} + {b}_{k}^{(2)})\\
    {C}^{(2)}_{xyk}[t] &= (\textbf{W} * \textbf{H}^{(1)}[t])_{xyk}\\
    \tilde{{H}}^{(2)}_{xyk}[t] &= \zeta({\kappa}_{k}{H}^{(1)}_{xyk}[t] + {\beta}_{k}{C}^{(2)}_{xyk}[t] + {\omega}_{k}{H}^{(1)}_{xyk}[t]{C}^{(2)}_{xyk}[t])\label{h2}\\
    {H}^{(2)}_{xyk}[t] &= \eta_{t}({H}^{(2)}_{xyk}[t-1]({1} - {G}^{(2)}_{xyk}[t]) + \tilde{{H}}^{(2)}_{xyk}[t]{G}^{(2)}_{xyk}[t] )
\end{align}


The mix $\textbf{G}^{(2)}[t]$ is calculated by convolving $\textbf{U}^{(2)}[t]$ with $\textbf{H}^{(1)}[t]$, followed by the addition of $\textbf{b}^{(2)}$. The activity $\textbf{C}^{(2)}[t]$ represents the excitation of horizontal connections onto the newly-computed $\textbf{H}^{(1)}[t]$. Linear and quadratic contributions of horizontal interactions at this stage are controlled by the $k$-dimensional parameters $\boldsymbol{\kappa}$, $\boldsymbol{\omega}$, and $\boldsymbol{\beta}$. The parameters $\boldsymbol{\kappa}$ and $\boldsymbol{\omega}$ control the linear and quadratic contributions of horizontal connections to $\tilde{\textbf{H}}^{(2)}[t]$. The parameter $\boldsymbol{\beta}$ is a gain applied to $\textbf{C}^{(2)}[t]$, giving $\textbf{W}$ an additional degree of freedom in expressing this excitation. With this full suite of interactions, the hGRU can in principle implement both a linear and a quadratic form of excitation (\ie to assess self-similarity), each of which play specific computational roles in perception~\citep{Martin2004-ud}. Note that the inclusion of $\textbf{H}^{(1)}[t]$ in Eq.~\ref{h2} functions as a ``peephole'' connection between it and $\tilde{\textbf{H}}^{(2)}[t]$. Finally, the mix $\textbf{G}^{(2)}$ integrates the candidate $\tilde{\textbf{H}}^{(2)}_{t}$ with $\textbf{H}^{(2)}_{t}$. The learnable $T$-dimensional parameter $\boldsymbol{\eta}$, which we refer to as a time-gain, helps control unstable gradients during training. This time-gain modulates $\textbf{H}^{(2)}_{t}$ with the scalar, $\eta_{t}$, which as we show in our experiments below improves model performance.

\section{The Pathfinder challenge}

We evaluated the limits of feedforward and recurrent architectures on the ``Pathfinder challenge'', a synthetic visual task inspired by cognitive psychology~\citep{Houtkamp2010-te}. The task, depicted in Fig.~\ref{fig:task}b, involves detecting whether two circles are connected by a path. This is made more difficult by allowing target paths to curve and introducing multiple shorter unconnected ``distractor'' paths. The Pathfinder challenge involves three separate datasets, for which the length of paths and distractors are parametrically increased. This challenge therefore screens models for their effectiveness in detecting complex long-range spatial relationships in cluttered scenes.



\paragraph{Stimulus design}

Pathfinder images were generated by placing oriented ``paddles'' on a canvas to form dashed paths. Each image contained two paths made of a fixed number of paddles and multiple distractors made of one third as many paddles. Positive examples were generated by placing two circles at the ends of a single path (Fig.~\ref{fig:task}b, left) and negative examples by placing one circle at the end of each of the paths (Fig.~\ref{fig:task}b, right). The paths were curved and variably shaped, with the possible number of shapes exponential to the path length. The Pathfinder challenge consisted of three datasets, in which path and distractor length was successively increased, and with them, the overall task difficulty. These datasets had path lengths of 6, 9 and 14 paddles, and each contained 1,000,000 unique images of 150$\times$150 pixels. See Supp. Material for a detailed description of the stimulus generation procedure.

\begin{figure}[t]
\begin{center}\vspace{5mm}
  \includegraphics[width=1\linewidth]{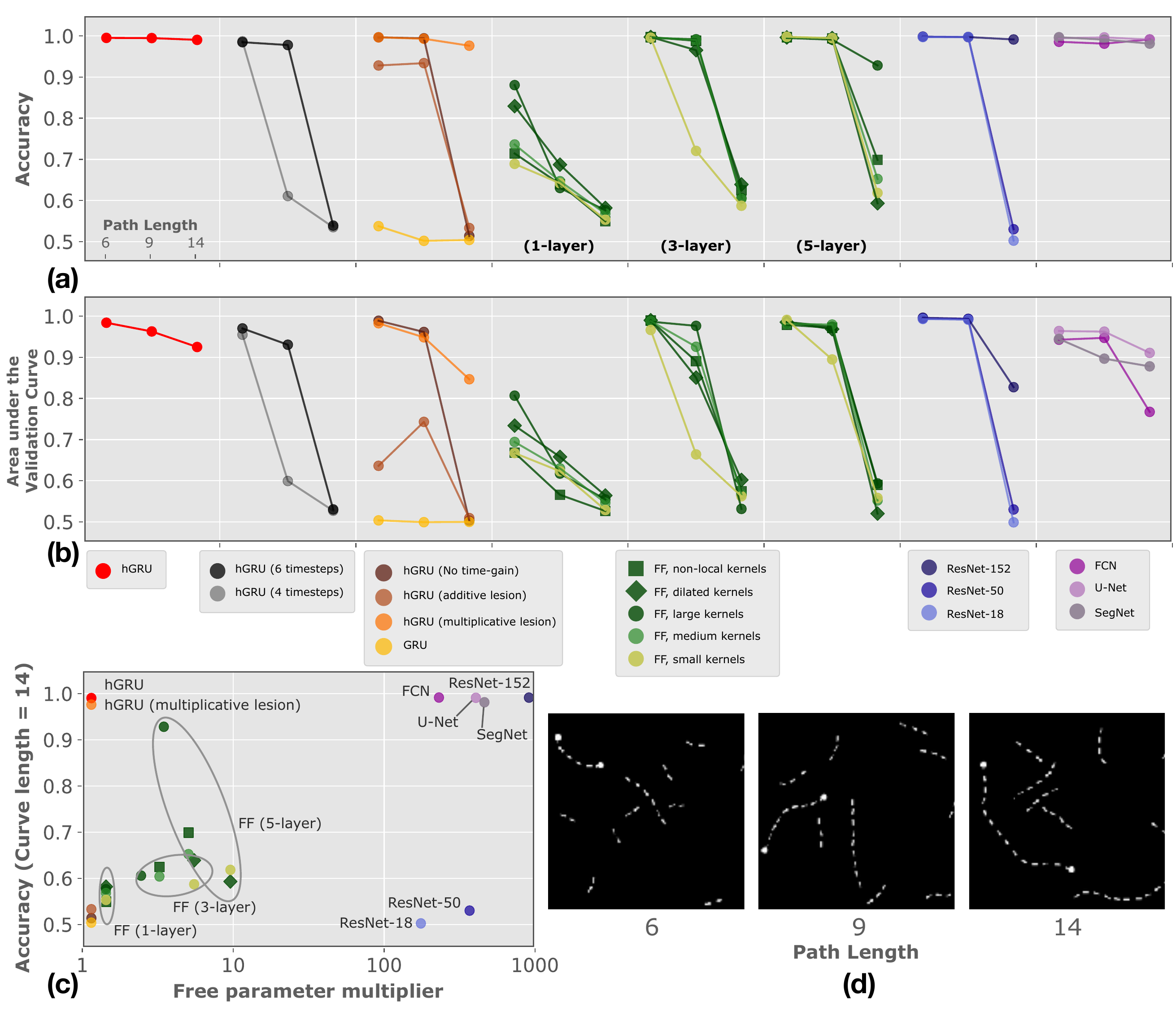}
  \caption{The hGRU efficiently learns long-range spatial dependencies that otherwise strain feedforward architectures. \textbf{(a)} Model accuracy is plotted for the three Pathfinder challenge datasets, which featured paths of 6- 9- and 14-paddles. Each panel depicts the accuracy of a different model class after training on for each pathfinder dataset (see Supp. Material for additional models). Only the hGRU and state-of-the-art models for classification (the two right-most panels) approached perfect accuracy on each dataset. \textbf{(b)} Measuring the area under the learning curve (ALC) of each model (mean accuracy) demonstrates that the rate of learning achieved by the hGRU across the Pathfinder challenge is only rivaled by the U-Net architecture (far right). \textbf{(c)} The hGRU is significantly more parameter-efficient than feedforward models at the Pathfinder challenge, with its closest competitors needing at least 200$\times$ the number of parameters to match its performance. The x-axis shows the number of parameters in each model versus the hGRU, as a multiple of the latter. The y-axis depicts model accuracy on the 14-length Pathfinder dataset. \textbf{(d)} Pathfinder challenge exemplars of different path lengths (all are positive examples).}\label{pathfinder}
\end{center}
\end{figure}

\paragraph{Model implementation}
We performed a large-scale analysis of the effectiveness of feedforward and recurrent computations on the Pathfinder challenge. We controlled for the effects of idiosyncratic model specifications by using a standard architecture, consisting of ``input'', ``feature extraction'', and ``readout'' processing stages. Swapping different feedforward or recurrent layers into the feature extraction stage let us measure the relative effectiveness of each on the challenge. All models except for state-of-the-art ``residual networks'' (ResNets)~\citep{He2016-es} and per-pixel prediction architectures were embedded in this architecture, and these exceptions are detailed below. See Supp. Material for a detailed description of the input and readout stage. 
Models were trained on each Pathfinder challenge dataset (Fig.~\ref{pathfinder}d), with 90\% of the images used for training (900,000) and the remainder for testing (100,000). We measured model performance in two ways. First, as the accuracy on test images. Second, as the ``area under the learning curve'' (ALC), or mean accuracy on the test set evaluated after every 1000 batches of training, which summarized the rate at which a model learned the task. Accuracy and ALC were taken from the model that achieved the highest accuracy across 5 separate runs of model training. All models were trained for two epochs except for the ResNets, which were trained for four. Model training procedures are detailed in Supp. Material.

\paragraph{Recurrent models}
We tested 6 different recurrent layers in the feature extraction stage of the standard architecture: hGRUs with 8, 6, and 4-timesteps of processing; a GRU; and hGRUs with lesions applied to parameters controlling linear or quadratic horizontal interactions. Both the GRU and lesioned versions of the hGRU ran for 8 timesteps. These layers had 15$\times$15 horizontal connection kernels ($W$) with an equal number of channels as their input layer (25 channels). 

We observed 3 overarching trends: First, each model's performance monotonically decreased, or ``strained'', as path length increased. Increasing path length reduced model accuracy (Fig.~\ref{pathfinder}a), and increased the number of batches it took to learn a task (Fig.~\ref{pathfinder}b). Second, the 8-timestep hGRU was more effective than any other recurrent model, and it outperformed each of its lesioned variants as well as a standard GRU. Notably, this hGRU was strained the \emph{least} by the Pathfinder challenge out of all tested models, with a negligible drop in accuracy as path length increased. This finding highlights the effectiveness of the hGRU for processing long-range spatial dependencies, and how the dynamics implemented by its linear and quadratic horizontal interactions are important. Third, hGRU performance monotonically decreased with processing time. This revealed a minimum number of timesteps that the hGRU needed to solve each Pathfinder dataset: 4 for the length-6 condition, 6 for the length-9 condition, and 8 for the length-14 condition (first vs. second columns in Fig.~\ref{pathfinder}a). Such time-dependency in the Pathfinder task is consistent with the accuracy-reaction-time tradeoff found in humans as the distance between endpoints of a curve increases~\citep{Houtkamp2010-te}.


\paragraph{Feedforward models}
We screened an array of feedforward models on the Pathfinder challenge. Model performance revealed the importance of kernel size \vs kernel width, model depth, and feedforward operations for incorporating additional scene context for solving Pathfinder. Model construction began by embedding the feature extraction stage of the standard model with kernels of one of three different sizes: 10$\times$10, 15$\times$15, or 20$\times$20. These are referred to as small, medium, and large kernel models (Fig.~\ref{pathfinder}). To control for the effect of network capacity on performance, the number of kernels given to each model was varied so that the number of parameters in each model configuration was equal to each other and the hGRU (36, 16, and 9 kernels). We also tested two other feedforward models that featured candidate operations for incorporating contextual information into local convolutional activities. One version used (2-pixel) dilated convolutions, which involves applying a stride to the kernel before convolving the input~\citep{Long2015-ou,Yu2016-bh}, and has been found useful for many computer vision problems~\citep{Wang2017-sr,Hamaguchi2017-bc,Chen2018-tq}. The other version applied a non-local operation to convolutional activities~\citep{Wang2017-mh}, which can introduce (non-recurrent) interactions between units in a layer. These operations were incorporated into the first feature extraction layer of the medium kernel (15$\times$15 filter) model described above. We also considered deeper versions of each of the above ``1-layer'' models (referring to the depth of the feature extraction stage), stacking them to build 3- and 5-layer versions. This yielded a total of 15 different feedforward models.

Without exception, the performance of each feedforward model was significantly strained by the Pathfinder challenge. The magnitude of this straining was well predicted by model depth and size, and operations for incorporating additional contextual information made no discernible difference to the overall pattern of results. The 1-layer models were most effective on the 6-length Pathfinder dataset, but were unable to do better than chance on the remaining conditions. Increasing model capacity to 3 layers rescued the performance of all but the small kernel model on the 9-length Pathfinder dataset, but even then did little to improve performance on the 14-length dataset. Of the 5-layer models, only the large kernel configuration came close to solving the 14-length dataset. The ALC of this model, however, demonstrates that its rate of learning was slow, especially compared to the hGRU (Fig.~\ref{pathfinder}b). The failures of these feedforward models is all the more striking when considering that each had between 1$\times$ and 10$\times$ the number of parameters as the hGRU (Fig.~\ref{pathfinder}c, compare the red and green markers).

\paragraph{Residual networks}
We reasoned that if the performance of feedforward models on the Pathfinder challenge is a function of model depth, then state-of-the-art networks for object recognition with many times the number of layers should easily solve the challenge. We tested this possibility by training ResNets with 18, 50, and 152 layers on the Pathfinder challenge. Each model was trained ``from scratch'' with standard weight initialization \citep{He2015-lf}, and given additional epochs of training (4) to learn the task because of their large number of parameters. However, even with this additional training, only the deepest 152-layer ResNet was able to solve the challenge (Fig.~\ref{pathfinder}a). Even so, the 152-layer ResNet was less efficient at learning the 14-length dataset than the hGRU (Fig.~\ref{pathfinder}b), and achieved its performance with nearly 1000$\times$ as many parameters (Fig.~\ref{pathfinder}c; see Supp. Material for additional ResNet experiments).

\paragraph{Per-pixel prediction models}
We considered the possibility that CNN architectures for per-pixel prediction tasks, such as contour detection and segmentation, might be better suited to the Pathfinder challenge than those designed for classification. We therefore tested three representative per-pixel prediction models: the fully-convolutional network (FCN), the skip-connection U-Net, and the unpooling SegNet. These models used an encoder/decoder style architecture, which was followed by the readout processing stage of the standard architecture described above to make them suitable for Pathfinder classification. Encoders were the VGG16~\citep{Simonyan2014-jd}, and each model was trained from scratch with Xavier initialized weights. Like the ResNets, these models were given 4 epochs of training to accommodate their large number of parameters.

The fully-convolutional network (FCN) architecture is one of the first successful uses of CNNs for per-pixel prediction~\citep{Long2015-ou,Papandreou2015-qo,Xie2017-or,Maninis2017-ta}. Decoders in these models use ``1$\times$1'' convolutions to combine upsampled activity maps from several layers of the encoder. We created an FCN model which applied this procedure to the last layer of each of the 5 VGG16-convolution blocks. These activity maps were upsampled by learnable kernels, which were initialized with weights for bilinear interpolation. In contrast to the feedforward models discussed above, the FCN successfully learned all conditions in the Pathfinder challenge (Fig.~\ref{pathfinder}a, purple circle). It did so less efficiently than the hGRU, however, with a lower ALC score on the 14-length dataset and 200$\times$ as many free parameters (Fig.~\ref{pathfinder}b). 


Another approach to per-pixel prediction uses ``skip connections'' to connect specific layers of a model’s encoder to its decoder. This approach was first described in~\citep{Long2015-ou} as a method for more effectively merging coarse-layer information into a model's decoder, and later extended to the U-Net~\citep{Ronneberger2015-sk}. We implemented a version of the U-Net architecture that had a VGG16 encoder and a decoder. The decoder consisted of 5 randomly initialized and learned upsampling layers, which had additive connections to the final convolutional layer in each of the encoder's VGG16 blocks. Using standard VGG16 nomenclature to define one of these connections, this meant that ``conv 4\_3'' activity from the encoder was added to the second upsampled activity map in the decoder. The U-Net was on par with the hGRU and the FCN at solving the Pathfinder challenge. It was also nearly as efficient as the hGRU in doing so (Fig.~\ref{pathfinder}b), but used over 350$\times$ as many parameters as the hGRU (Fig.~\ref{pathfinder}c; see Supp. Materials for additional U-Net experiments). 

Unpooling models eliminate the need for feature map upsampling by routing decoded activities to the locations of the winning max-pooling units derived from the encoder. Unpooling is also a leading approach for a variety of dense per-pixel prediction tasks, including segmentation, which is exemplified by SegNet~\citep{Badrinarayanan2017-hm}. We tested a SegNet on the Pathfinder challenge. This model has a decoder that mirrors its encoder, with unpooling operations replacing its pooling. The SegNet achieved high accuracy on each of the Pathfinder datasets, but was less efficient at learning them than the hGRU, with worse ALC scores across the challenge (Fig.~\ref{pathfinder}b). The SegNet also featured the second-most parameters of any model tested, which was 400$\times$ more than the hGRU.


\section{Explaining biological horizontal connections with the hGRU}

Statistical image analysis studies have suggested that cortical patterns of horizontal connections, commonly referred to as ``association fields'', may reflect the geometric regularities of oriented elements present in natural scenes~\cite{Ben-Shahar2004-jo}. Because the hGRU is designed to capture such spatial regularities, we investigated whether it learned patterns of horizontal connections that resemble these association fields. We visualized the horizontal kernels learned by the hGRU to solve tasks (Fig. ~\ref{fig:human}). When trained on the the Pathfinder challenge, hGRU kernels resembled the dominant patterns of horizontal connectivity in visual cortex. Prominent among these patterns are (1) the antagonistic near-excitatory vs. far-inhibitory surround organization also found in the visual cortex~\citep{Shushruth2013-cd}; (2) the association field, with collinear excitation and orthogonal inhibition~\citep{Stettler2002-hb, Rockland1983-ki}; and (3) other higher-order surround computations~\citep{Tanaka2009-my}. We also visualized these patterns after training the hGRU to detect contours in the naturalistic BSDS500 image dataset~\citep{Arbelaez2011-lw}. These horizontal kernels took on similar patterns of connectivity, but with far more definition and regularity, suggesting that the hGRU learns best from natural scene statistics.


How well does the hGRU explain human psychophysics data? We tested this by recreating the synthetic contour detection dataset used in~\citep{Li2002-bq}. This task had human participants detect a contour formed by co-linearly aligned paddles in an array of randomly oriented distractors. Multiple versions of the task were created by varying the distance between paddles in the contour (5 conditions). Contour detection accuracy of the hGRU was recorded on each of dataset for comparison with participants in~\citep{Li2002-bq}, whose responses were digitally extracted with WebPlotDigitizer from~\citep{Li2002-bq} and averaged (N=2). Plotting hGRU accuracy against the reported ``detection score'' revealed that increasing inter-paddle distance caused similar performance straining for both (Fig.~\ref{fig:human}b).





\vspace{-2mm}
\section{Discussion}\vspace{-2mm}\label{discussion}

The present study demonstrates that long-range spatial dependencies generally strain CNNs, with only very deep and state-of-the-art networks overcoming the visual variability introduced by long paths in the Pathfinder challenge. Although feedforward networks are generally effective at learning and detecting relatively rigid objects shown in well-defined poses, these models tend towards a brute-force solution when tasked with the recognition of less constrained structures, such as a path connecting two distant locations. This study adds to a body of work highlighting examples of routine visual tasks where CNNs fall short of human performance~\citep{Ricci2018-br, Szegedy2013-yi, Nguyen2015-in, Volokitin2017-he, Ellis2015-fr}.

We demonstrate a solution to the Pathfinder challenge inspired by neuroscience. The hGRU leverages computational principles of visual cortical circuits to learn complex spatial interactions between units. For the Pathfinder challenge, this translates into an ability to represent the elements forming an extended path while ignoring surrounding clutter. We find that the hGRU can reliably detect paths of any tested length or form using just a single layer. This contrasts sharply with the successful state-of-the-art feedforward alternatives, which used much deeper architectures and orders of magnitude more parameters to achieve similar success. The key mechanisms underlying the hGRU's performance are well known in computational neuroscience~\citep{Grossberg1985-ui,Series2003-mg,Zhaoping2011-dn,Shushruth2012-dv,Rubin2015-ws,Mely2018-bc}. However, these mechanisms have been typically overlooked in computer vision (but see~\citep{George2017-ae} for a successful vision model using horizontal connections and shown to break text-based CAPTCHAs).

We also found that hGRU performance on the Pathfinder challenge is a function of the \emph{amount of time} it was given for processing. This finding suggests that it concurrently expands the facilitative influence of one end of a target curve to the other while suppressing the influence of distractors. The performance of the hGRU on the Pathfinder challenge captures the iterative nature of computations used by our visual system during similar tasks~\citep{Roelfsema2011-wg} -- exhibiting a comparable tradeoff between performance and processing-time~\citep{Houtkamp2010-te}.  Visual cortex is replete with association fields that are thought to underlie perceptual grouping~\citep{Field1993-yp,Gilbert2013-hb}. Theoretical models suggest that patterns of horizontal connections reflect the statistics of natural scenes, and here too we find that horizontal kernels in the hGRU learned from natural scenes resemble cortical patterns of horizontal connectivity, including association fields and the paired near-excitatory / far-inhibitory surrounds that may be responsible for many contextual illusions~\citep{Shushruth2013-cd,Mely2018-bc}. The horizontal connections learned by the hGRU reproduce another aspect of human behavior, in which the saliency of a straight contour decreases as the distance between its paddles increases. This sheds light on a possible relationship between horizontal connections and saliency computation. 

In summary, this work diagnoses a computational deficiency of feedforward networks, and introduces a biologically-inspired solution that can be easily incorporated into existing deep learning architectures. The weights and patterns of behavior learned by the hGRU appear consistent with those associated with the visual cortex, demonstrating its potential for establishing novel connections between machine learning, cognitive science, and neuroscience.

\subsubsection*{Acknowledgments}

This research was supported by NSF early career award [grant number IIS-1252951] and DARPA young faculty award [grant number YFA N66001-14-1-4037]. Additional support was provided by the Carney Institute for Brain Science and the Center for Computation and Visualization (CCV) at Brown University. 



\clearpage


\setcounter{figure}{0}

\makeatletter 
\renewcommand{\thefigure}{S\@arabic\c@figure}
\makeatother

\setcounter{table}{0}

\makeatletter 
\renewcommand{\thetable}{S\@arabic\c@table}
\makeatother

\setcounter{page}{1}

\begin{centering}
\huge{\bf{Supplementary Material}} \\ \vspace{1cm}
\end{centering}

\renewcommand{\thefigure}{S\arabic{figure}}

%
%





\section*{Deriving the hGRU from the contextual neural circuit model}
From Eq. 1, we obtain the following by rearranging the decay term:
\begin{align}
\dot{H}_{xyk}^{(1)} &= -\eta^{-1}\epsilon^2 H_{xyk}^{(1)} + \eta^{-1} \lb \xi X_{xyk} - (\alpha H_{xyk}^{(1)} + \mu)C_{xyk}^{(1)}\rb_+\nonumber\\
\dot{H}_{xyk}^{(2)} &= -\eta^{-1}\sigma^2 H_{xyk}^{(2)} + \tau^{-1} \lb\gamma C_{xyk}^{(2)}\rb_+,
\end{align}
where $[\cdot]_+=\max(\cdot,0)$ is the ReLU function.

Now, we would like to discretize this equation using Euler's method. We first simplify the above equation by choosing $\eta=\tau$ and $\sigma=\epsilon$:
\begin{align*}
\dot{H}_{xyk}^{(1)} &= -\eta^{-1}\epsilon^2 H_{xyk}^{(1)} + \eta^{-1} \lb \xi X_{xyk} - (\alpha H_{xyk}^{(1)} + \mu)C_{xyk}^{(1)}\rb_+\nonumber\\
\dot{H}_{xyk}^{(2)} &= -\eta^{-1}\epsilon^2 H_{xyk}^{(2)} + \eta^{-1} \lb \gamma C_{xyk}^{(2)}\rb_+,
\end{align*}
Now, apply Euler's method with timestep $h$. This gives the following difference equation:
\begin{align*}
H_{xyk}^{(1)}[t]&=H_{xyk}^{(1)}[t-1] + h \lp -\eta^{-1}\epsilon^2 H_{xyk}^{(1)}[t-1] + \eta^{-1} \lb \xi X_{xyk}[t-1] - (\alpha H_{xyk}^{(1)}[t-1] + \mu)C_{xyk}^{(1)}[t]\rb_+\rp\\
H_{xyk}^{(2)}[t]&=H_{xyk}^{(2)}[t-1] + h \lp -\eta^{-1}\epsilon^2 H_{xyk}^{(2)}[t-1] + \eta^{-1}\lb\gamma C_{xyk}^{(2)}[t]\rb_+\rp.
\end{align*}
Distributing $h$, we get
\begin{align*}
H_{xyk}^{(1)}[t]&=H_{xyk}^{(1)}[t-1] - h \frac{\epsilon^2}{\eta} H_{xyk}^{(1)}[t-1] + h\eta^{-1} \lb \xi X_{xyk}[t-1] - (\alpha H_{xyk}^{(1)}[t-1] + \mu)C_{xyk}^{(1)}[t]\rb_+\\
H_{xyk}^{(2)}[t]&=H_{xyk}^{(2)}[t-1] - h \frac{\epsilon^2}{\eta} H_{xyk}^{(2)}[t-1] + h \eta^{-1}\lb\gamma C_{xyk}^{(2)}[t]\rb_+.
\end{align*}
Now, notice that if we choose $h=\frac{\eta}{\epsilon^2}$, the first two terms on the RHS of each line will cancel:
\begin{align}
H_{xyk}^{(1)}[t]&=\epsilon^{-2}\lb \xi X_{xyk}[t-1] - (\alpha H_{xyk}^{(1)}[t-1] + \mu)C_{xyk}^{(1)}[t]\rb_+\nonumber\\
H_{xyk}^{(2)}[t]&=\epsilon^{-2}\lb\gamma C_{xyk}^{(2)}[t]\rb_+.
\label{droovid}
\end{align}

We now have a discrete-time approximation of the initial dynamical system, shown in Eq.~\ref{euler}. Note that this model can also be thought of as a convolutional RNN with ReLU nonlinearity. Because RNNs are difficult to train in practice, the state of the art is to incorporate learned ``gates'' to manage the flow of information over time \citep{Tallec2018-hg}.

\begin{align}
\begin{split}
G^{(1)}_{xyk}[t] &= \sigma\lp X_{xyk} + (\mathbf{U}^{(1)} * \mathbf{H}^{(2)}[n-1])_{xyk} + {b}_k^{(1)}\rp\\
G^{(1)}_{xyk}[t] &= \sigma\lp X_{xyk} + (\mathbf{U}^{(2)} * \mathbf{H}^{(1)}[n])_{xyk} + {b}_k^{(2)}\rp,
\label{gates}
\end{split}
\end{align}

where $\sigma$ is a squashing pointwise nonlinearity, $\mathbf{U}^{(\cdot)}$ are convolution kernels, and $b^{(\cdot)}$ are bias vectors. Applied to Eq.~\ref{droovid}, these gates integrate past information with new computations more flexibility than the Euler integration above:
\begin{align}
\begin{split}
H^{(1)}_{xyk}[t]
&= \epsilon^{-2} G^{(1)}_{xyk}[t] \lb \xi X_{xyk}[t-1] - (\alpha H_{xyk}^{(1)}[t-1] + \mu)C_{xyk}^{(1)}[t]\rb_+\\
&\qquad+(1-G^{(1)}_{xyk}[t])H^{(1)}_{xyk}[t-1]\phantom{\sum_1}\\
H^{(2)}_{xyk}[t]&=\epsilon^{-2} G^{(2)}_{xyk}[t] \lb\gamma C_{xyk}^{(2)}[t]\rb_+ + (1-G^{(2)}_{xyk}) H^{(2)}_{xyk}[t-1]
\end{split}
\end{align}

\begin{figure}[t!]
    \centering
    \includegraphics[width=0.99\linewidth]{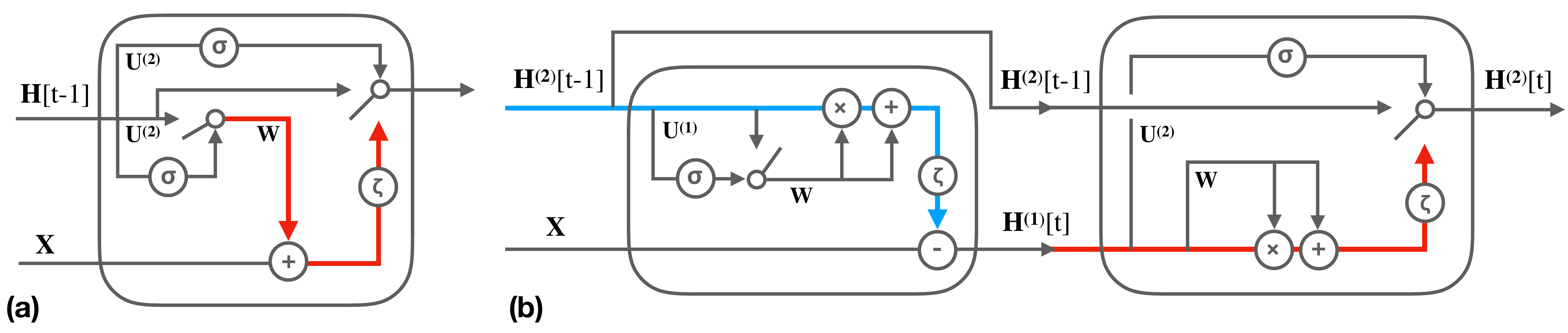}
    \caption{Circuit diagrams of the gated recurrent unit (GRU) and the hGRU showing the exitatory (red) and inhibitory (blue) flow of information. (a) A reference diagram of the GRU, demonstrating how the kernels ${U^{(1)}}$ and $U^{(2)}$ act as the ``gain'' and ``mix'' on the circuit's computations: selecting feature channels from $H[t-1]$ to process on the current timestep, and determining how to mix a candidate $\tilde H[t]$ with $H[t-1]$ to produce $H[t]$. Candidate activities are calculated by applying a hyperbolic tangent (tanh) to a linear combination of the feedforward drive $X$ and the convolution of the kernel $W$ with the new ``gain''-modulated hidden state. (b) The same gain and mix are used in the hGRU in a slightly different configuration. Note that in the hGRU, the mix is a function of $\textbf{H}^{(1)}[t]$ instead of $\textbf{H}^{(2)}[t-1]$. The separate processing stages of the hGRU, inspired by the recurrent neural circuit of \citep{Mely2018-bc}, are critical for learning the Pathfinder challenge.}
    \label{fig:diagram_si}
\end{figure}

This defines the simplest gated version of the RNN described in Eq.~\ref{droovid}. As discussed in the main text, we modified the model in several other ways to improve its effectiveness on the visual tasks examined here: (1) We found that using the GRU-style ``gain'' (eq 3.) to control the first stage of horizontal interactions was more effective than using ``mixing'' gates in both stages. (2) We found it effective to replace the $\epsilon^{-2}$ scaling by the learned per-timestep parameter $\boldsymbol{\eta}$, which can be thought of as a restricted application of batch normalization to the RNN's hidden activity to control saturating activities stabilize training~\citep{Cooijmans2017-bf}. (3) We included both linear and quadratic forms of excitation in $\mathbf{H}^{(2)}$ for symmetry with $\mathbf{H}^{(1)}$. Enabling the model to spread excitation via both linear and quadratic propagation is potentially useful for propagating activity based on either first- (e.g., contrast strength) or second-order (e.g., self-similarity) statistical regularities. The linear and quadratic terms are scaled per-feature by the learnable parameters $\boldsymbol{\alpha}$ and $\boldsymbol{\mu}$ when calculating $\mathbf{H}^{(1)}$, and $ \boldsymbol{\kappa}$ and $\boldsymbol{\omega}$, and $\boldsymbol{\omega}$ in $\mathbf{H}^{(2)}$. See \ref{fig:diagram_si} for a comparison of circuit diagrams for the GRU \vs the hGRU.

\section*{The Pathfinder challenge}

\paragraph{Overview}

The main goal of the Pathfinder challenge is to assess the ability of a computer vision system to determine whether two distant locations in an image, marked by while circles, are connected by a path. The stimulus dataset consists of binary images, each containing two white circles, two white `target' paths, and multiple distractor paths, placed on a black background. All images used in our experiments are rendered on a 150$\times$150 canvas. The task is inspired by~\cite{Roelfsema2011-wg}.

\begin{figure}[t]
\begin{center}
  \includegraphics[width=1\linewidth]{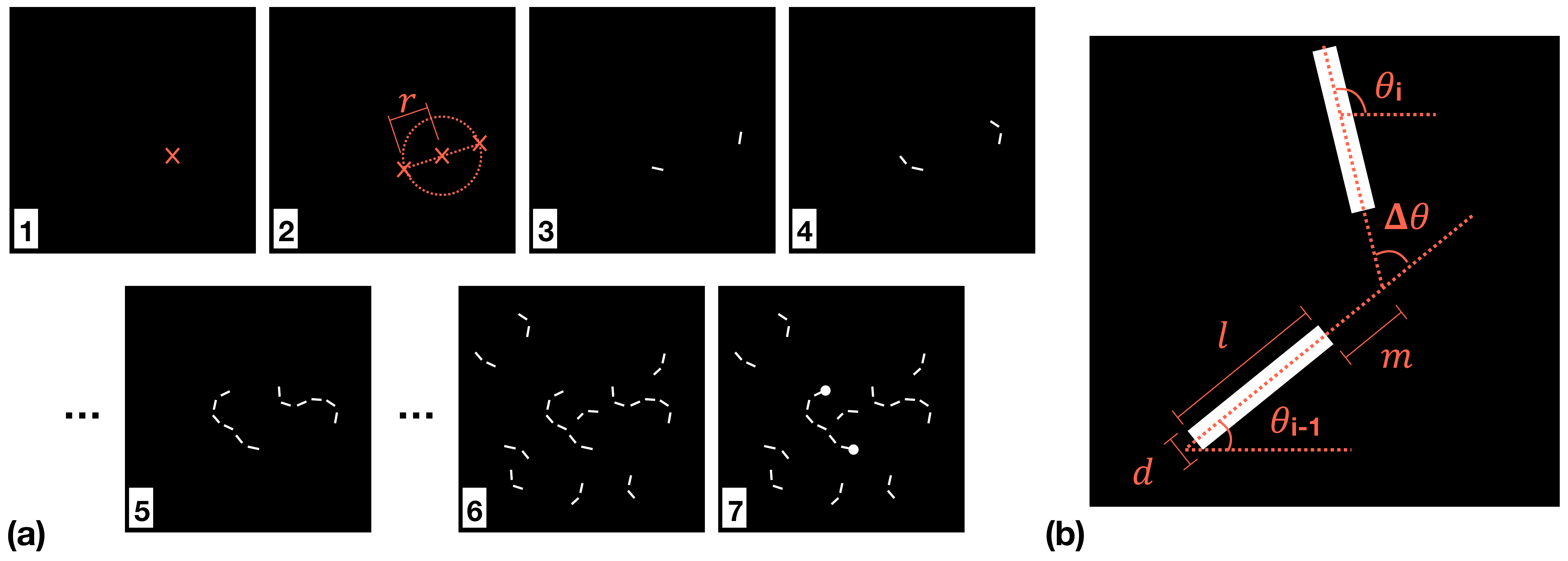}
  \caption{Image generator algorithm. (a) Depiction of image generation shown in seven steps. (b) Depiction of the geometric constraints for trailing paddle placement during path generation.}\label{generator_fig}

\end{center}
\end{figure}

Each image is generated by first sampling two target paths and then rendering multiple distractor paths. Each path consists of $k$ co-circularly arranged identically shaped ``paddles'' of length $l$ and thickness $d$ (Fig.~\ref{generator_fig}b). A paddle is a white oriented bar, characterized by the position of its center and its orientation, $\theta$. Each path is generated inductively, namely, by first sampling the position of its ``seed paddle'' which serves as the end of the path and then iteratively adding new ``trailing paddles'' next to the seed paddle or the last trailing paddle. The high-level overview of the image generation algorithm is depicted in Fig.~\ref{generator_fig} and the list of image parameters in Table~\ref{generator_fig}.

\paragraph{Positioning target paths} 

The first stage of generating target paths involves randomly sampling the position of an invisible circle of radius $r$ (Fig.~\ref{generator_fig}a, step 1) within an empty image. Then, a randomly oriented line pivoted at the center of the circle makes two intersections with the circle. These intersections serve as the positions of two target paths' seed paddles (Fig.~\ref{generator_fig}a, step 2). This constraint ensures that target paths are always located within a sufficient proximity to each other. Without this constraint, target paths can be separated by an arbitrary distance, which may allow a model to make classification decision solely based on the distance between the white markers, for the markers will tend to be located much farther apart in negative examples. Then, two randomly oriented paddles are rendered at each of the intersections (Fig.~\ref{generator_fig}a, step 3), serving as seed paddles of the target paths.

\paragraph{Growing a path} 

Trailing paddles are sequentially added to the end of each target path (Fig.~\ref{generator_fig}a, step 4 and 5). Each trailing paddle is added by first sampling its orientation, $\theta_i$, according to the probability distribution defined by the angle formed between the new and the previous trailing paddle, $\Delta \theta = \theta_{i} - \theta_{i-1}$:

\vspace{-5mm}\begin{align}
    P(\theta_i) = \frac{1}{Z}\max(\cos{(c\Delta \theta)}, 0)
\end{align}\vspace{-4mm}
\begin{align}\vspace{-4mm}
    \text{where } Z = \int_{-\frac{\pi}{2}}^{\frac{\pi}{2}}\cos(\theta)d\theta
    = 2
\end{align}

Note that the continuity parameter $c$ determines the overall rigidity of a path by constraining the possible range of orientations of each new trailing paddle. With the sampled orientation, the new trailing paddle's position is determined such that the line extending from the trailing paddle intersects with the line extending from the new paddle $m$ pixels away from the end of the two paddles while also forming an angle $\Delta\theta$ (Fig.~\ref{generator_fig}b). Note that the parameter $m$ determines the margin between adjacent paddles in a path. Because we do not allow paddles to cross or touch, we add new paddles to the two target paths in alternation to ensure that the shapes of one path does not restrict the shape of the other. The total number of target path paddles, or simply length of target paths, is denoted by the parameter $k$.

Additional distractor paths consisting of $\frac{k}{3}$ paddles are added to the image. Their generation process is identical to that of target paths, except that a seed paddle's location is independently sampled. The total number of distractor paths is chosen so that the total number of paddles in an image is 150. The usage of two target paths as well as distractor paths prevents any `short-cut' solution to this problem, such as detecting a loose end of a path or a lone white circle as a local diagnostic cue for classification. Additionally, this design allows us to ensure that positive and negative images are indistinguishable in terms of the way paddles are arranged. This forces a model to perform classification solely based on the connectedness of the white circles.

As mentioned, the generator does not allow paddles to be too close to each other or make any contact. If a newly sampled paddle is making contact with or not separated from other paddles by at least $m$ pixels, the generator rejects it and re-samples a paddle. In practice, this is done by applying circular dilation on the paddles using a kernel of radius $m$ and checking if the dilated images of paddles has any overlap.

\paragraph{Shape variability}

Because adding an additional paddle to a path multiplicatively increases the number of possible shapes of the path, path shape variability is exponential to the total number of paddles in a path, $k$. Similarly, the continuity parameter $c$ controls path shape variability in an exponential manner because it scales the range of possible angles formed between every pair of adjacent paddles in a path. In our experiment, we vary path length $k$ to examine how model performance changes as the shape variability of a path in a dataset increases. All images in the Pathfinder challenge are generated using Python and OpenCV.

\begin{table}[]
\centering
\caption{Image parameters in the Pathfinder challenge.}
\label{my-label}
\begin{tabular}{ll}
Notation & Definition                                                      \\ \hline
$r$        & Radius of a circle                                              \\
$k$        & Target path length, or the number of paddles in a target path \\
$l$        & Paddle length, in pixels                                        \\
$d$        & Paddle thickness, in pixels                                     \\
$m$        & Inter-paddle margin, in pixels                                  \\
$c$        & Continuity                                                     
\end{tabular}
\label{params}
\end{table}

\section*{The Synthetic Contours Dataset}

In this section, we shall explain the motivation and construction procedure for the Synthetic Contours Dataset. We created the Synthetic Contours Dataset (SCD) to compare an hGRU pre-trained on the BSDS500~\citep{Arbelaez2011-lw} for contour detection against human behavioral data as described in~\citep{Li2002-bq}. 

\paragraph{Dataset generation} We generated the SCD by following closely, and extending the dataset construction procedure described in~\citep{Li2002-bq}. SCD enables us to parametrically control the position, orientation, and relative spacing of generated contours, resulting in 28 million images with practically infinite variability along these directions.  All images in the SCD have a resolution of 256$\times$256 pixels, spanning a radius of 4 visual degrees. Each contour image is rendered by placing ``paddles'' of length 0.1 visual degrees within uniformly spaced cells on a master grid. This master grid covers the entire image with a height and a width of 8 visual degrees. Each cell in the grid spans a height and width of 0.25 visual degrees. A total of 32 paddles are placed along every row/column of the master grid.

A contour of length $l$ is generated in the master grid by filling $l$ neighboring grid cells on any grid-diagonal, and each paddle connects the diagonally opposite points in their respective cells. This aligns the paddles in a collinear manner to form a contour of length $l$. In order to ensure that the majority of the contour stays within the image, its center is positioned within a maximum distance from the center of the image. This distance from the image center, known as eccentricity, is denoted as $e$. Once the contour is generated, all remaining unoccupied cells are filled with paddles each with an orientation $\theta \in [0,2\pi]$ sampled from a uniform distribution.

\begin{figure}[h]
    \centering
    \includegraphics[width=0.99\linewidth]{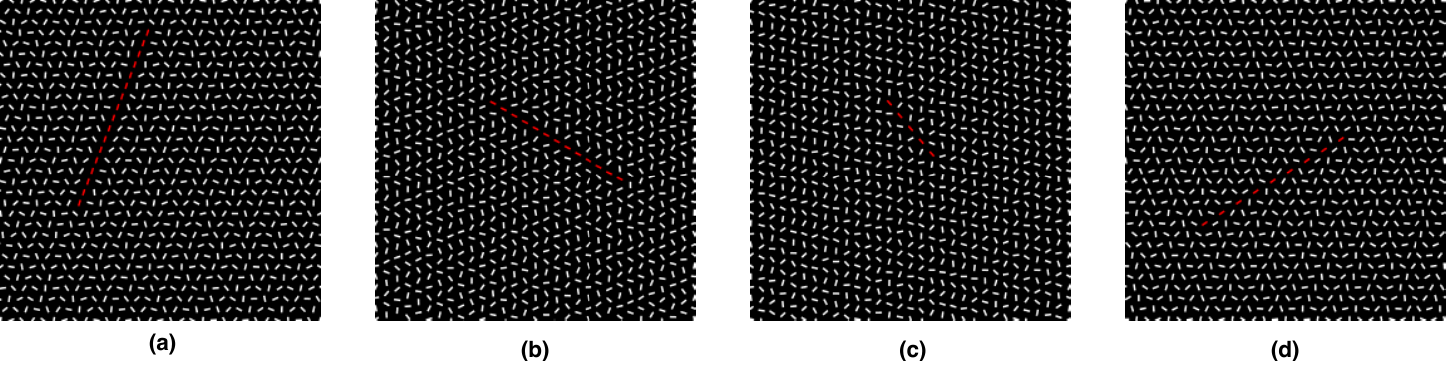}
    \caption{Examples of images from the SCD. Unlike in the dataset, contours in the figure are highlighted with a different color for the reader's convenience. (a) Contour with \textit{SF}=-0.6, \textit{l}=17, closest relative spacing. (b) Contour with \textit{SF}=-0.4, \textit{l}=14. (c) Contour with \textit{SF}=0.2, \textit{l}=9. (d) Contour with \textit{SF}=0.6, \textit{l}=9, farthest relative spacing.}    
    \label{fig:gilbert_si}
\end{figure}

\paragraph{Parametric dataset variability}
Contours in SCD are formed by regulating the dataset variability in 2 different directions: (1) contour length and (2) relative spacing. We denote the number of paddles present in a contour as its length. Each image in the SCD contains a contour with a length of either 5, 9, 14 or 17 paddles. The distance between neighboring paddles that form a contour is denoted as the relative spacing. While we vary the relative spacing, the global spacing between distractor paddles is maintained constant. Following ~\citep{Li2002-bq}, we apply a shear operation to the generated image by a shear factor \textit{SF}, which controls how close or far the neighboring contour paddles are positioned. We generated images with 5 different relative-spacing conditions by varying SF between -0.6 and +0.6. A comparison between human performance and model performance on the task of contour detection can be found in Fig.~\ref{fig:human}. Example figures with different combinations of contour length and relative spacing are demonstrated in Fig.~\ref{fig:gilbert_si}. All the images in SCD were generated using the Python Psychophysics toolbox, Psychopy~\citep{Peirce2007-ey}. 

\section*{Model architecture details}

All of the recurrent models and feedforward models (shown in shades of green in Fig.~\ref{pathfinder}a in the main text) tested in our study share the `input stage' as the preprocessing module and the `readout stage' as the classification module, which allows us to more directly compare the effectiveness of different architectures (called `feature extraction' stage) at solving the Pathfinder challenge. Unless otherwise specified, all feedforward model weights were Xavier initialized~\citep{Glorot2010-fa}. Recurrent model kernels and parameters were also initialized with Xavier, while their biases were chronos-initialized \citep{Tallec2018-hg}. Hidden states of te recurrent models were also randomly initialized. Implementations of the hGRU can be found at \url{https://github.com/serre-lab/hgru_share}, and datasets are available upon request.

Standard cross entropy was used to compute loss. Models were trained with gradient descent using the Adaptive Moment Estimation (Adam) optimizer~\citep{Kingma2014-js} with base learning rate of $10^{-3}$ on batches of 32 images. Our experiments are conducted using TensorFlow~\citep{Abadi2016-co}.

\paragraph{Input stage}

The input stage consists of a convolutional layer with 25 kernels of size 7$\times$7. The kernels are initialized as Gabor filters at 12 orientations and 2 phases, plus a radially symmetric difference-of-Gaussian filter. Filter activity undergoes a pointwise squaring non-linearity before being passed to the model-specific feature extraction stage. Note that none of the off-the-shelf models (residual networks and per-pixel prediction models) use this input stage. 

\paragraph{Readout stage}

The readout stage takes the output from the final layer of the feature extraction stage and computes a two-dimensional likelihood vector for deriving a decision on an image from the Pathfinder challenge. The readout stage contained three parts: (1) a 1x1 convolutional filter that transforms the multi-channel output map from the feature extraction stage to a two-channel map, each corresponding to the positive/negative class. (2) A batch-normalized~\citep{Ioffe2015-ug} global max pool, whose output represents in each channel the maximum activation value across space. (3) A linear classifier which maps the two-dimensional feature vector into a class decision. The configuration of this readout stage allows us to compare the accuracy between models designed for per-image prediction with those designed for per-pixel prediction. Note that residual networks do not use a readout stage as they are proposed as standalone image classification architectures.

\paragraph{Additional recurrent models}
We tested three other kinds of recurrent models that were not discussed in the main text. These were the hGRU with batchnorm (hGRU-BN), the hGRU with a nonnegative pointwise nonlinearty (hGRU-ReLU), and a two-layer GRU (GRU-2L).

The hGRU formulation in the main text incorporates a highly constrained form of the normalization discussed in \citep{Cooijmans2017-bf}, in which a learned scalar modulates activity at every timestep to help control unstable gradients. We found that incorporating the full batchnorm formulation of \citep{Cooijmans2017-bf} in the hGRU also performed well on the Pathfinder challenge. We implemented the resulting hGRU-BN by adding batchnorm to multiple computations in the circuit. While it is possible to share batchnorm parameters across timesteps, the model performs better when they are not shared. In this formulation, ${H}^{(1)}[t]_{xyk}$ is calculated as
\begin{align}
\begin{split}
    \textbf{G}^{(1)}[t] &= \sigma(\mathrm{BN}(\textbf{U}^{(1)} * \textbf{H}^{(2)}[t-1]))\\
    {C}^{(1)}[t]_{xyk} &= \mathrm{BN}(\textbf{W} * (\textbf{G}^{(1)}[t] \odot \textbf{H}^{(2)}[t-1]))_{xyk}\\
    {H}^{(1)}_{xyk}[t] &= \zeta({X}_{xyk} - {C}^{(1)}_{xyk}[t](\alpha_{k} {H}^{(2)}_{xyk}[t-1] +\mu_{k}))\label{inhibition_2}
\end{split}
\end{align}
where
\vspace{-3mm}\begin{align*}
\mathrm { BN } ( \mathbf { h } ; \boldsymbol{\delta} , \boldsymbol{\nu} ) = \boldsymbol{\nu} + \boldsymbol{\delta} \odot \frac { \mathbf { h } - \widehat { \mathbb { E } } [ \mathbf { h } ] } { \sqrt { \widehat { \operatorname { Var } } [ \mathbf { h } ] + \epsilon } }
\end{align*}

Where $\textbf{h} \in \mathbb { R } ^ { d }$ is the vector of preactivations being normalized by batchnorm and $\boldsymbol{\delta}$, $\boldsymbol{\nu} \in \mathbb { R } ^ { d }$ are parameters that control the standard deviation and mean of the normalized activities, $\epsilon$ is a regularization hyperparameter, and $\odot$ is elementwise multiplication. As in the hGRU described in the main text, this updated $\textbf{H}^{(1)}[t]$ is next used to calculate $\textbf{H}^{(2)}[t]$.

\begin{align}\vspace{-4mm}
    \textbf{G}^{(2)}[t] &= \sigma(\mathrm{BN}(\textbf{U}^{(2)} * \textbf{H}^{(1)}[t])))\\
    {C}^{(2)}[t]_{xyk} &= \mathrm{BN}(\textbf{W} * \textbf{H}^{(1)}[t])_{xyk}\\
    \tilde{{H}}^{(2)}_{xyk}[t] &= \zeta({\kappa}_{k}{H}^{(1)}_{xyk}[t] + {\beta}_{k}{C}^{(2)}_{xyk}[t] + {\omega}_{k}{H}^{(1)}_{xyk}[t]{C}^{(2)}_{xyk}[t])\\
    {H}^{(2)}_{xyk}[t] &= {H}^{(2)}_{xyk}[t-1]({1} - {G}^{(2)}_{xyk}[t]) + \tilde{{H}}^{(2)}_{xyk}[t]{G}^{(2)}_{xyk}[t]
\end{align}

\begin{figure}[t!]
\begin{center}    \includegraphics[width=0.99\linewidth]{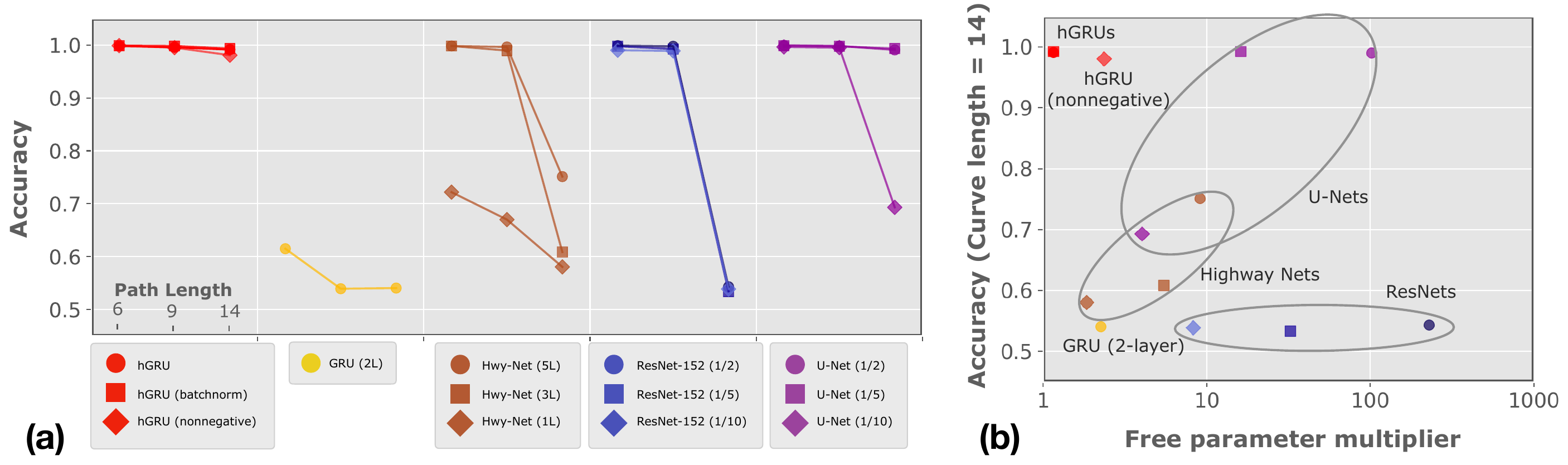}
    \caption{A comparison of the hGRU and additional control models on the Pathfinder challenge. These models are hGRUs with additional normalization stages (``batchnorm'') or constraints (``nonnegative''); a GRU with two layers of processing; feedforward models with highway network layers; 152-layer ResNets restricted to a fraction of the total parameters of the original model; and U-Net models restricted to a fraction of the total parameters of the original model. (a) While all versions of the hGRU have high accuracy on all three Pathfinder challenge datasets, only the U-Net limited to $\frac{1}{2}$ or $\frac{1}{5}$ the total number parameters performs similarly. (b) Model accuracy is plotted as a function of the multiple of parameters each has \wrt the hGRU (\ie hGRU is 1 and all models with more parameters are $>$ 1). This reveals a model complexity and performance gap between the hGRU and other classes of feedforward models. Per-pixel prediction models, such as the U-Net, perform relatively well on the Pathfinder challenge but need many more parameters than the hGRU to do so: note the difference in performance between the U-Net ($\frac{1}{10}$) and the hGRU. Highway nets seem to follow a similar performance trajectory, and 152-layer ResNets struggle with the task without their full set of parameters (see Fig.~\ref{pathfinder} in the main text).}\label{fig:accuracy_si}
\end{center}
\end{figure}

Performance of the hGRU-BN can be found in \ref{fig:accuracy_si} as ``hGRU (batchnorm)''. In addition to its ability to solve the Pathfinder challenge, the hGRU-BN is far more stable during training than the original hGRU formulation when its pointwise nonlinearity $\zeta$ is substituted for nonlinearities that are not squashing. This allowed us to build a version of the hGRU where the pointwise nonlinearity $\zeta$ is set to linear rectification (ReLU), which constrains $\textbf{H}^{(1)}$ to inhibition and $\textbf{H}^{(2)}$ to excitation as is done in the original model by \citep{Mely2018-bc}. Also similar to \citep{Mely2018-bc} this model used separate kernels $\textbf{W}^{I}$ and $\textbf{W}^{E}$ to calculate facilitative \vs suppressive horizontal interactions at $\textbf{H}^{(1)}$ and $\textbf{H}^{(2)}$. This model solved the pathfinder challenge, and is denoted in Fig. \ref{fig:accuracy_si} as ``hGRU (nonnegative)''.

We also tested whether the hGRU's ability to solve the Pathfinder challenge was merely a function of its two processing stages. We developed a GRU with two layers of convolution (with separate kernels) to calculate its candidate hidden state. Although this ``GRU (2L)'' model performed better than the typical one-layer version on the 6-length pathfinder, it was unable to solve the other versions of the problem (Fig. \ref{fig:accuracy_si}).

\paragraph{Additional models}
We extended our screening of feedforward architectures on the Pathfinder challenge to include three new model classes. The first combines highway network modules \citep{Srivastava2015-pj} with the input- and readout-stages that were used to screen feedforward model operations in the main text. As with those models, we tested the highway network in one-, three-, and five-layer configurations. The highway network models performed similarly to the ``large-kernel'' configuration of feedforward models screened in the main text, solving the 6- and 9-length pathfinder tasks, but straining on the 14-length one (see ``Hwy-Net'' in Fig. \ref{fig:accuracy_si}).

We also tested ``constrained'' versions of our top-performing feedforward and per-pixel prediction models, the ResNet-152 and the U-Net. We created new versions of these models that had $\frac{1}{2}$, $\frac{1}{5}th$, or $\frac{1}{10}th$ the total number of parameters. The resulting ResNet models solved the 6- and 9-length pathfinder datasets were unsuccessful on the 14-length pathfinder. In constrast, the constrained U-Nets performed far better, with only the U-Net~$\frac{1}{10}$ strained by the 14-length pathfinder. Plotting model performance as a function of their multiple of parameters \vs the hGRU reveals a performance shelf for these feedforward models that depend on depth to expand their receptive fields and solve tasks such as the Pathfinder challenge. This strategy is less efficient than the hGRU's ability to form interactions between units \textit{at a single processing layer}. The model with the closest performance and number of parameters as the hGRU in this comparison is the $\frac{1}{5}$ U-Net, which still needed over an order of magnitude more free parameters to solve the challenge.

\section*{Training on natural images and comparisons to human data}

\begin{figure}[t!]
\begin{center}
  \includegraphics[width=1\linewidth]{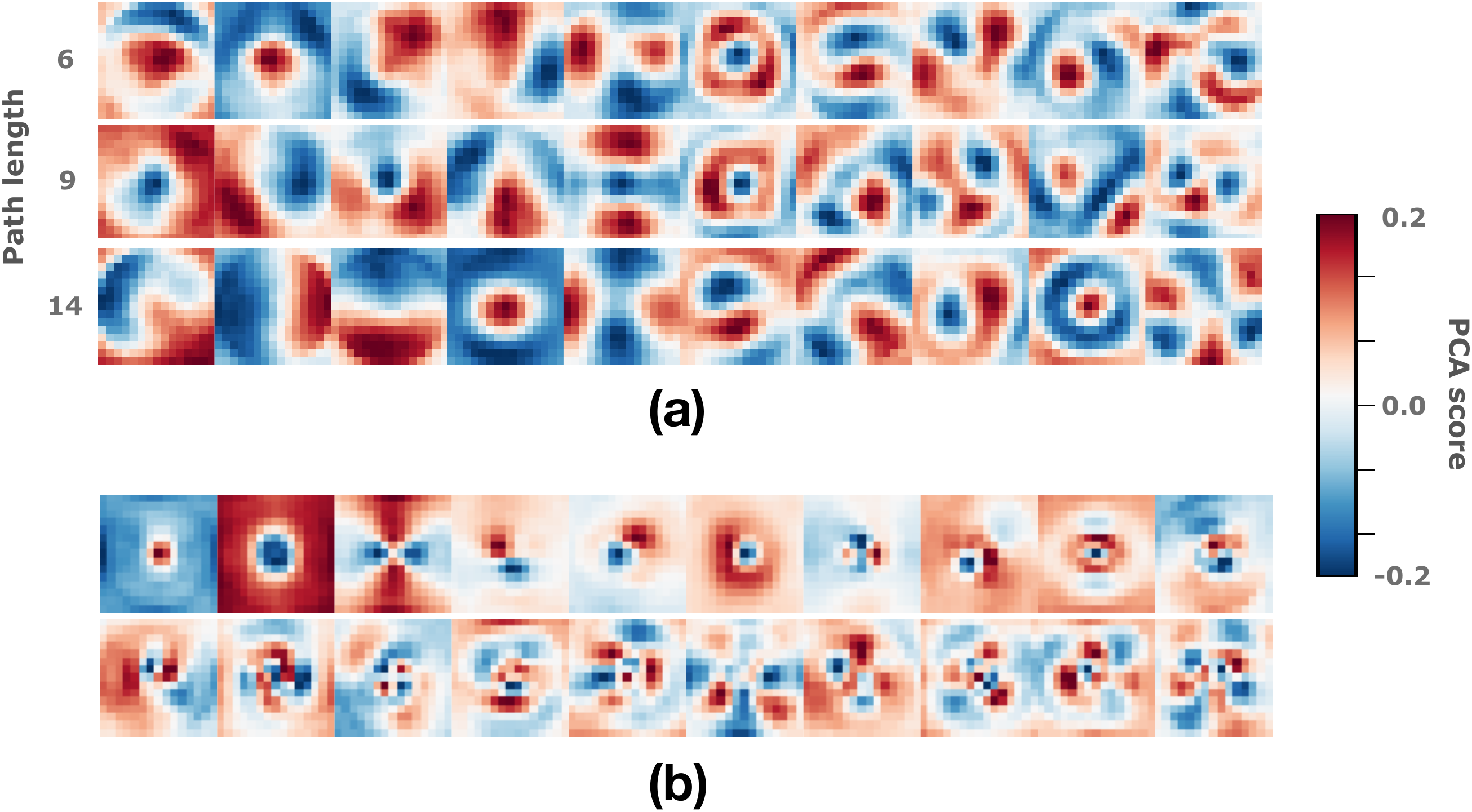}
  \caption{hGRU eigenconnectivity on when trained on the Pathfinder challenge and contour detection in natural scenes. (a) Eigenconnectivity of hGRU models trained on 6-, 9-, and 14-length Pathfinder datasets. (b) Eigenconnectivity of an hGRU on the BSDS500 contour detection dataset. The exhibited kernels contain patterns that are similar to those observed in visual cortex.}
\end{center}
\label{kernels}
\end{figure}

\paragraph{Visualizing horizontal connections}
Theoretical models of visual cortex suggest that patterns of horizontal connections reflect statistical regularities of oriented features in natural scenes~\citep{Ben-Shahar2004-jo}. Here too we find that horizontal kernels in the hGRU learned from both synthetic images from the Pathfinder challenge and natural scenes resemble cortical patterns of horizontal connectivity, including association fields and the paired near-excitatory and far-inhibitory surrounds (Fig.~\ref{kernels}b).

We visualized hGRU horizontal connectivity in models through a two-step procedure that aligned kernels into a common reference orientation (90 degrees), and then used PCA to identify and denoise common patterns. For each input channel of the hGRU's horizontal kernel, we rotated the kernels in the opposite orientation of the previous layer's filter used to compute their feedforward drive. For example, for hGRUs trained on the Pathfinder challenge, the feedforward drive came from kernels initialized with oriented gabors. Here, a gabor oriented around 90 degrees would result in its horizontal kernels being rotated -90 degrees. After normalizing the orientation of all horizontal kernels, they were mean centered and then passed through PCA to visualize their ``eigenconnectivity'' patterns in Fig.~\ref{kernels}. In other words, these eigenconnectivities correspond to the spatial associations learned by horizontal kernels, aggregated over all orientation channels. Eigenconnectivity patterns are sorted by their eigenvalues (80\% cumulative variance cutoff).


We plot hGRU eigenconnectivities learned from two tasks: the Pathfinder challenge (Fig.~\ref{kernels}a) and contour detection in natural images (Fig.~\ref{kernels}b). For the first task, we use models described in the main text trained on each of the Pathfinder datasets. For the second task, we trained a new version of the hGRU to detect contours in the BSDS500~\citep{Arbelaez2011-lw}. We used an 8-timestep with the first two blocks of PASCAL weight initialized convolutional layers from the VGG16 as its input processing stage. This model was trained for 1000 epochs on the 200 training images in the BSDS500, using data augmentations (random crops to 300$\times$300 pixels, random rotations of +/- 30$^{\circ}$, and random up/down/left/right flips), a per-pixel cross-entropy loss, and the same learning rate and optimizer as the models described above.

Pathfinder challenge eigenconnectivity is similar to canonical patterns of cortex, including the (1) the antagonistic near-excitatory vs. far-inhibitory surround organization of hypercolumns in the visual cortex~\citep{Shushruth2013-cd}; (2) the association field, with collinear excitation and orthogonal inhibition~\citep{Stettler2002-hb, Rockland1983-ki}; and (3) higher-order surround computations~\citep{Tanaka2009-my}.

Horizontal kernel eigenconnectivities are far cleaner and qualitatively more similar to purported canonical patterns of connectivity in cortex after training the hGRU to detect contours in natural images (compare Fig.~\ref{kernels}a and Fig.~\ref{kernels}b). We believe that such a difference at least in part results from the the kinds of image statistics that are useful for the respective tasks of Pathfinder \vs BSDS -- path integration \vs contour detection. The path integration task in the Pathfinder challenge can be solved by relying nearly exclusively on the co-linearity of neighboring paddles whereas contour detection in natural images might oftentimes require richer, sometimes non-directional, measures of local change.

\paragraph{hGRU explains human contour detection}

The patterns of horizontal connectivity such as association fields are thought to underlie human performance in contour detection~\citep{Gilbert2013-hb}. Given the presence of visually similar horizontal connectivity in the hGRU, we were motivated to test its ability to explain human psychophysics data during contour detection. We did this by recreating synthetic datasets from~\citep{Li2002-bq} that were used to measure human contour detection performance. The task involved detecting a contour of co-linearly aligned paddles in an array of randomly oriented distractor paddles. Different versions of the task were generated by varying inter-paddle distance within a contour (5 conditions).

We generated 1,000,000 unique 150$\times$150 pixel images for every condition (28 total). In each case, 90\% of images (900,000) were used for training and the remaining 10\% (100,000) for testing. We fine-tuned our BSDS-trained hGRU separately on each of these training datasets and recorded its accuracy on the corresponding test datasets for comparison with human observers.

\begin{figure}[t!]
\begin{center}
    \includegraphics[width=0.8\linewidth]{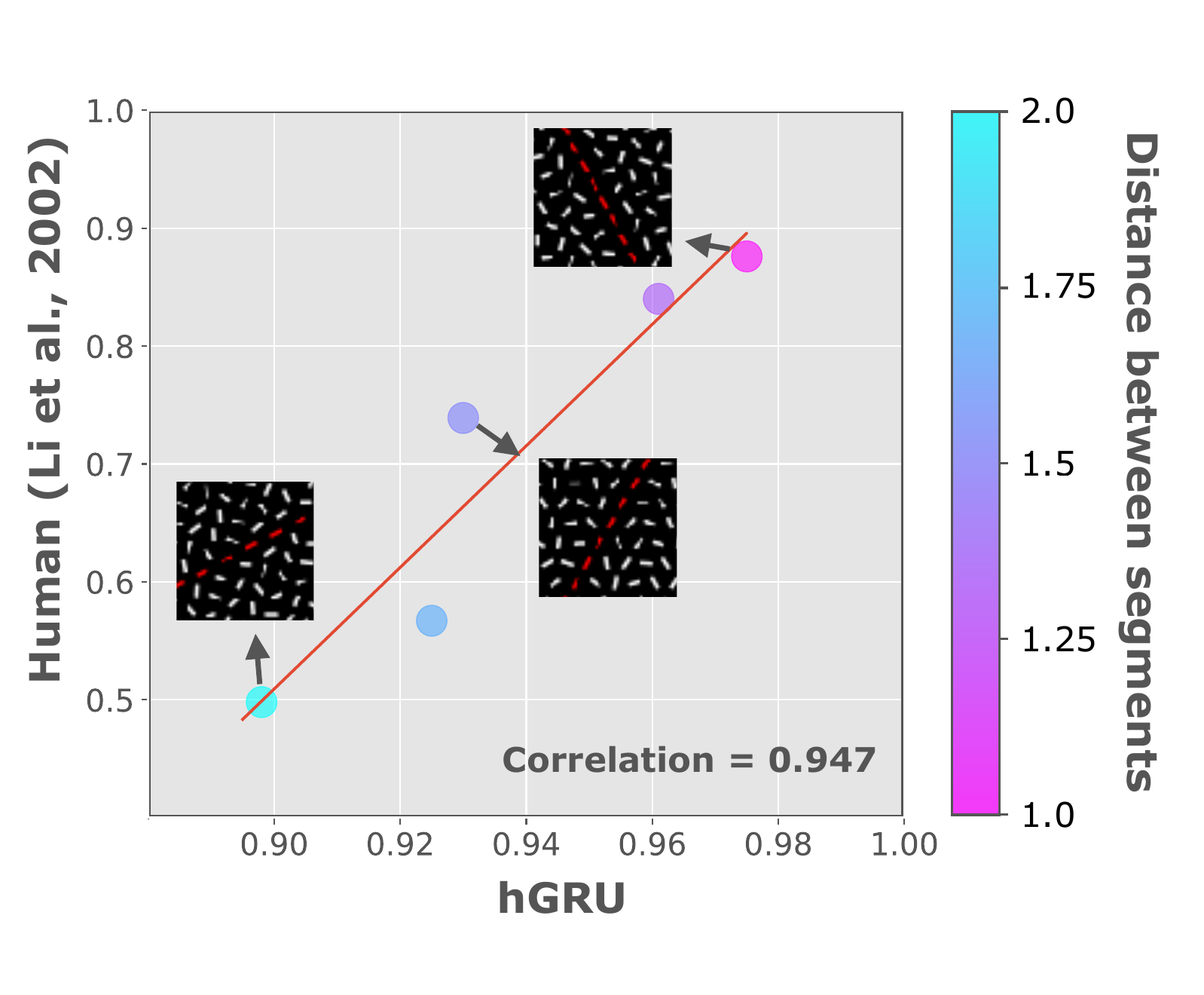}
   \caption{An hGRU trained on contour detection in natural images and then fine-tuned on a classic contour detection task performs similarly to human observers. While the hGRU is overall more accurate than humans, the performance of both is strained as the distance between paddles increases. Inlays show example contour stimuli with varying distance between their paddles (from left to right: larger-to-smaller gaps). Note that we use red to highlight contour stimuli for visualization, but it is absent during the task.}\label{fig:human}
\end{center}
\end{figure}

Although the hGRU was accurate at detecting paddles when the distance between paddles increased, this manipulation still strained network performance. We compared hGRU performance with human participants, whose responses were digitally extracted from~\citep{Li2002-bq} and averaged together for every experimental condition. Plotting the accuracy of the hGRU against the reported ``detection score'' of human observers revealed similar straining for both in response to increasing inter-paddle distance (Fig.~\ref{fig:human}).

\end{document}